\documentclass[letterpaper]{article} 
\usepackage[preprint]{aaai2027}  
\usepackage[hyphens]{url}  
\usepackage{graphicx} 
\usepackage{natbib}  
\usepackage{caption} 
\usepackage{booktabs}

\usepackage{amsmath}
\usepackage{amssymb}
\usepackage{bbding} 
\usepackage{algorithm}
\usepackage{algorithmic}
\usepackage{multirow}

\newcommand{\state}{s}

\newcommand{\hist}{\mathcal{H}}
\newcommand{\action}{a}
\newcommand{\chunk}{A}

\newcommand{\selector}{f_{\phi,\theta}^{\mathrm{sel}}}
\newcommand{\rank}{\rho}

\newcommand{\BCE}{\operatorname{BCE}}

\title{SparkVLA: Stop-Aware Hierarchical VLA with Adaptive Action Chunking for Long-Horizon Manipulation}
\author{
    Xunyao Lei\textsuperscript{1,*},
    Renjun Wu\textsuperscript{1,*},
    Tianlin Huo\textsuperscript{2},
    Xuesong Li\textsuperscript{1,\Envelope}
}
\affiliations{
    \textsuperscript{1}Beijing Institute of Technology\\
    \textsuperscript{2}Tsinghua University\\
    \textsuperscript{*}Equal contribution\quad
    \textsuperscript{\Envelope}Corresponding author\\
    Project page: \url{https://icr-lab.github.io/SparkVLA}
}

\begin{document}
\maketitle
\begin{abstract}
At every re-observation point in a hierarchical
Vision-Language-Action (VLA) system, two interface decisions must be
made: \emph{when} to terminate the current subtask and \emph{how far}
to execute the proposed action chunk.  These decisions are mutually
dependent---the optimal stopping point depends on what the executor
plans to do, while the optimal execution length depends on where the
subtask boundary lies---yet existing architectures evaluate them in
isolation, an asymmetry neither module can overcome alone.
We present \textbf{SparkVLA}, a stop-aware hierarchical VLA that
resolves this mutual dependency by formulating both decisions as a
single ranking: \textsc{Stop} competes against every action-prefix
length in a unified candidate set, and the system selects the
highest-scoring option, eliminating threshold tuning and requiring only
offline ordinal preferences.
An \emph{Anchor-Conditioned Context Encoding} module caches a
history-aware subtask anchor encoding onset-state memory and goal
semantics, guiding visual-token pruning toward task-relevant regions;
a \emph{Stop-Aware Action-Prefix Selection} head scores all candidates
via full self-attention at chunk boundaries for efficiency.
On RoboCerebra, SparkVLA achieves
\textbf{47.12\%}~success rate, surpassing the official hierarchical baseline by
\textbf{30.57\%} and the strongest reproducible method by 26.83\% Real-robot experiments on
multi-step tasks further validate these gains on physical hardware.
\end{abstract}


\section{Introduction}

Recent Vision-Language-Action (VLA) models~\cite{rt2, openvla, pi0, octo, pi05, rdt, groot, smolvla} have demonstrated
remarkable generalization on robotic manipulation, yet they
are predominantly trained on short trajectory segments and
struggle to compose multiple skills over long
horizons~\cite{hivla, robocerebra}.
Hierarchical VLA architectures~\cite{saycan, hivla, hirt, g0, mem, rmbench}
address this by pairing a high-level VLM planner with a low-level VLA
executor, achieving multi-step manipulation~\cite{mem, g0},
cross-embodiment transfer~\cite{robodual}, and improved long-horizon
robustness~\cite{rmbench}.
A systematic study~\cite{hivla} identifies the \emph{interface
mechanisms}---particularly the termination condition and execution
horizon---as the highest-leverage design choices.

Yet a critical interface decision remains under-explored: at every
re-observation boundary the system must simultaneously judge
\emph{when to terminate the current subtask} and \emph{how far to
execute the proposed action chunk}.  These two decisions jointly govern
closed-loop behavior---premature termination discards progress,
late termination causes overshooting, over-commitment leads to drift,
and under-commitment wastes replanning budget---yet existing approaches
treat each in isolation.
On the termination side, hierarchical systems employ binary
detectors~\cite{g0}, VLM-predicted lengths~\cite{hivla}, or progress
signals~\cite{palm}---all deciding \emph{whether} to stop without
access to the upcoming plan.  On the horizon side,
DEHP~\cite{dehp}, ACH~\cite{ach}, AQC~\cite{aqc}, and
AAC~\cite{aac} adaptively select how far to commit but cannot signal
subtask completion.
Crucially, these decisions are \emph{mutually dependent}: the optimal
stopping point depends on what the executor plans to do next, while the
optimal execution length depends on where the subtask boundary lies.
This mutual dependency cannot be resolved by independent
modules---it demands joint evaluation.

\begin{figure*}[t]
\centering
\includegraphics[width=0.90\textwidth]{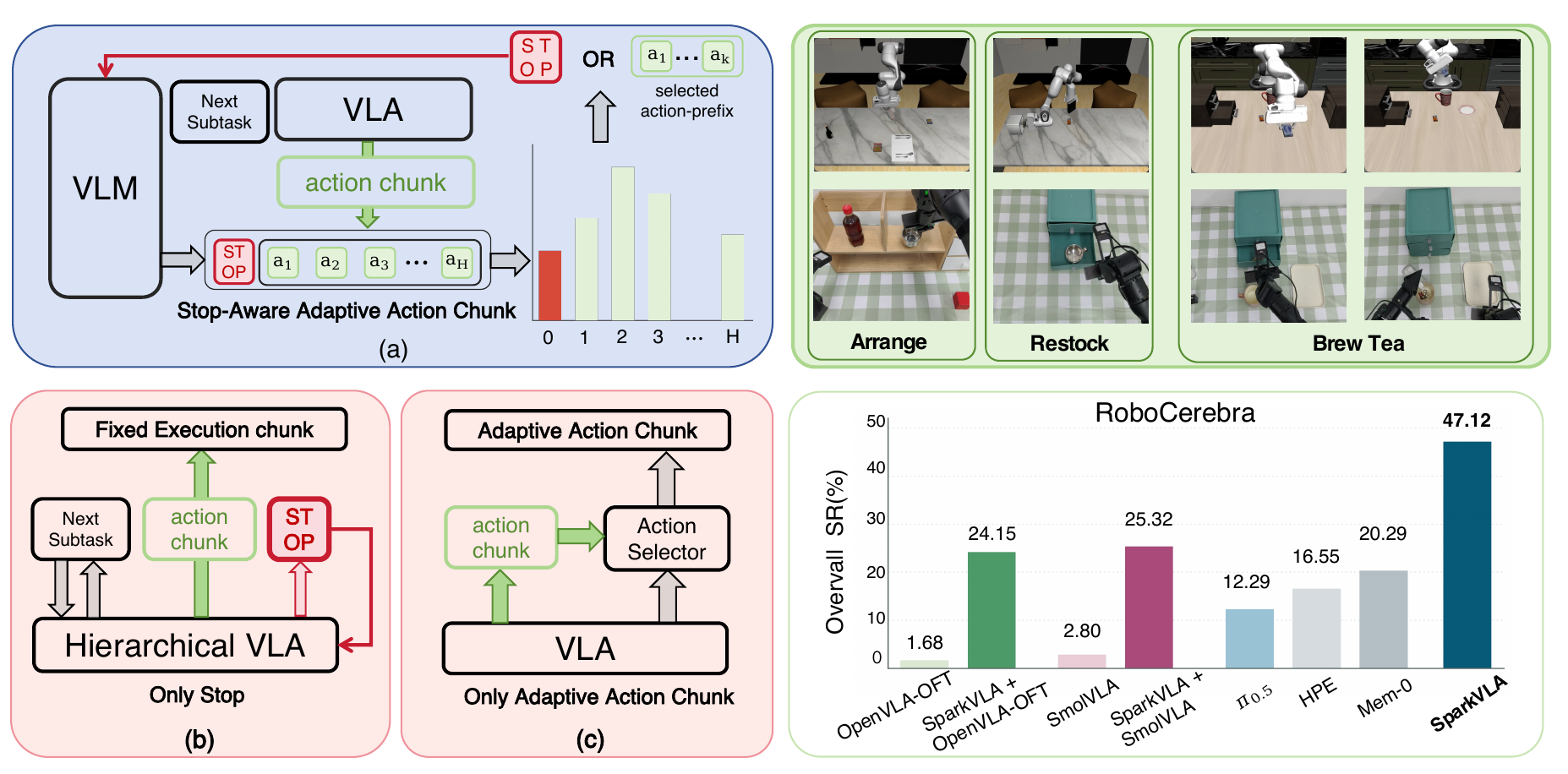}
\caption{Existing hierarchical VLAs either execute fixed-length chunks
with a separate stop detector~(b) or adapt the horizon without an
explicit termination signal~(c), addressing each decision in isolation.
SparkVLA~(a) unifies both into a single ranking over \textsc{Stop} and
all action-prefix lengths, enabling joint adaptive execution and
threshold-free subtask termination.}
\label{fig:teaser}
\end{figure*}

We observe that both decisions reduce to a common assessment---how well
the proposed chunk serves the current subtask from the present
state---and formulate a unified candidate set in which \textsc{Stop}
competes against every action-prefix length under the same scoring head
(Figure~\ref{fig:teaser}).  Because all candidates are scored within a
single shared representation, termination can leverage the content of
the proposed plan and horizon selection can leverage proximity to the
subtask boundary---precisely the mutual information that isolated
modules lack.  Both decisions then reduce to selecting the
highest-scoring candidate, eliminating threshold tuning and requiring
only ordinal supervision.

Building on this insight, we present SparkVLA, a stop-aware hierarchical
VLA built from a shared pretrained VLA: the high-level
branch retains the VLM backbone for subtask generation and hosts the
selector, while the low-level branch produces action chunks.
An \emph{Anchor-Conditioned Context Encoding} module caches a
history-aware subtask anchor encoding onset-state memory and goal
semantics, guiding visual-token pruning toward task-relevant regions.
A \emph{Stop-Aware Action-Prefix Selection} head then scores all
candidates via full self-attention and selects the highest-scoring
option, trained from offline ordinal preferences.  The selector operates only at chunk
boundaries, improving both speed and accuracy over per-step
monitoring.

On RoboCerebra, SparkVLA achieves 47.12\% success rate, outperforming the
official hierarchical baseline by 30.57\% and the strongest
reproducible method by 26.83\%.  Real-robot experiments on three
multi-step tasks validate these gains on physical hardware.

Our contributions are threefold:
\begin{itemize}
\item We formulate subtask termination and execution-horizon selection
  as a single ranking over a unified candidate set, resolving the mutual
  dependency without separate detectors, threshold tuning, or online
  reward signals.
\item We introduce \emph{Anchor-Conditioned Context Encoding} and
  \emph{Stop-Aware Action-Prefix Selection}, realizing this decision
  with a single scoring pass at chunk boundaries.
\item We demonstrate that SparkVLA's high-level system generalizes
  across different low-level executors  and transfers in both simulation and real-world platforms.
\end{itemize}
\begin{figure*}[t]
    \centering
    \includegraphics[width=\textwidth]{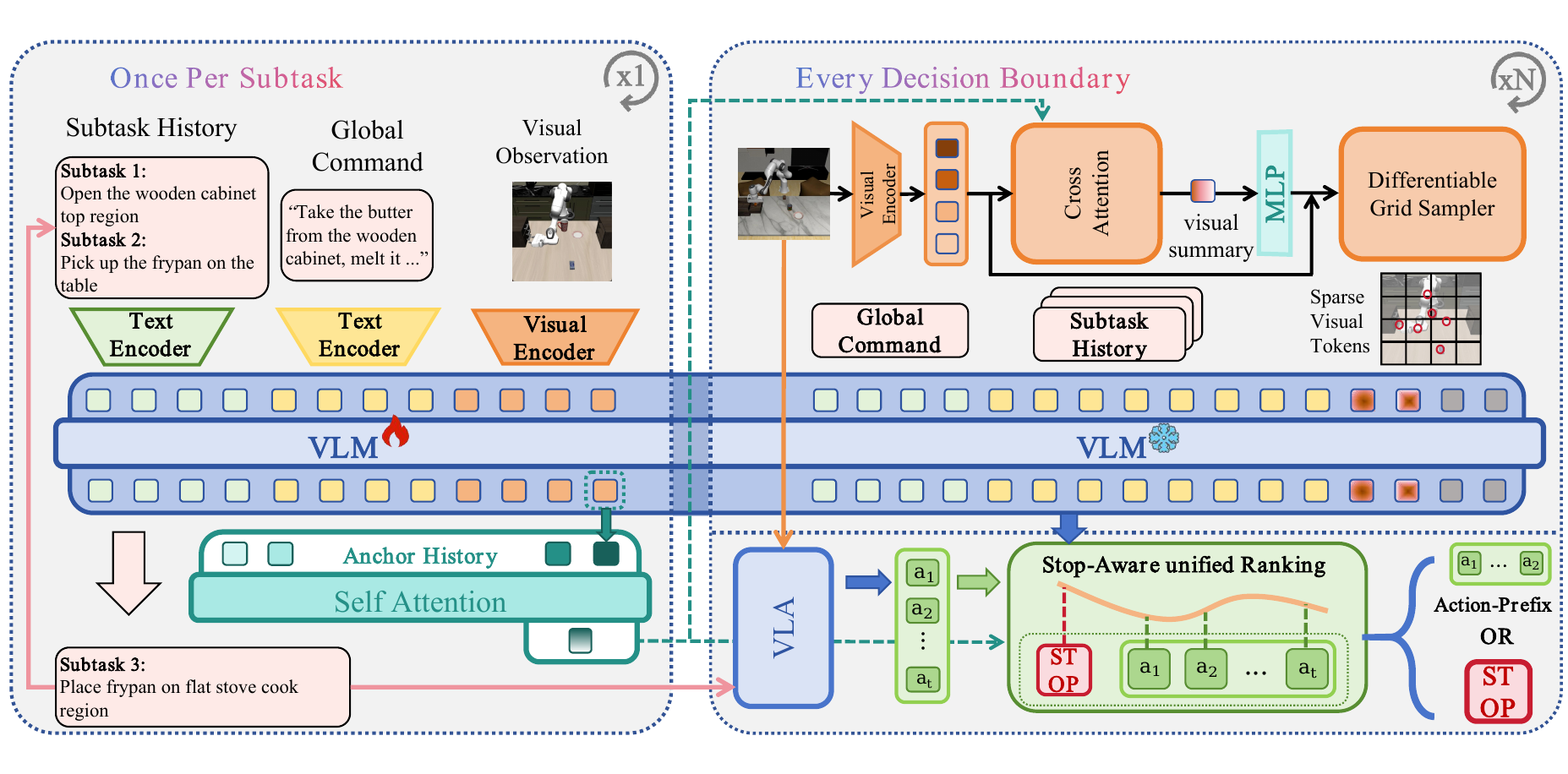}
    \caption{Overview of SparkVLA. At each subtask onset, the high-level
    planner autoregressively generates a subtask command, extracts a raw
    anchor from the penultimate layer, and fuses it with the anchor
    history pool via self-attention to form a history-aware anchor.
    At every decision boundary, anchor-conditioned context encoding
    extracts task-relevant visual features, while the stop-aware selector
    ranks \textsc{Stop} against all action-prefix candidates.}
    \label{fig:overview}
\end{figure*}

\section{Related Work}

\subsection{Hierarchical VLA Systems}

Hierarchical architectures for robotic manipulation separate high-level
planning from low-level execution to tackle long-horizon
tasks~\cite{hivla, hirt, g0, robodual, mem, rmbench}.
Representative systems pass latent features~\cite{hirt}, subtask-level
language commands~\cite{g0}, multi-scale memory~\cite{mem}, or
task-phase classifiers~\cite{rmbench} across the interface.
MEM~\cite{mem} combines short-term video with long-term language memory,
enabling tasks spanning up to fifteen minutes; PALM~\cite{palm}
integrates affordance reasoning and continuous progress signals for
subtask-aware execution.
A systematic study~\cite{hivla} unifies these designs under an
options-framework loop and identifies the termination condition as the
highest-leverage interface mechanism.
Despite this variety, all existing systems treat subtask termination as
an isolated binary decision---judging whether to stop without access to
the upcoming action plan or the execution horizon it implies.  SparkVLA
instead places termination inside a unified ranking over the proposed
chunk, enabling joint assessment with execution length.

\subsection{Adaptive Action Chunking}

Action chunking---predicting and committing to a short action sequence
before replanning---is standard in modern visuomotor
policies~\cite{aloha, diffusion_policy}, but a fixed
execution horizon creates a fundamental tradeoff between smoothness and
reactivity.
DEHP~\cite{dehp} trains a lightweight RL head to
predict execution length while keeping the base policy frozen.
ACH~\cite{ach} and AQC~\cite{aqc} estimate Q-values for all candidate
lengths in one forward pass, selecting the highest-value horizon without
requiring online interaction.
AAC~\cite{aac} uses action entropy as a training-free cue for adaptive
chunk-size selection at inference time.
These methods improve over fixed horizons but cannot signal subtask completion.  SparkVLA
bridges both lines by scoring \textsc{Stop} and all action-prefix
lengths under a shared head, trained from offline ordinal preferences
without online RL.
\section{Method}


\subsection{Problem Formulation}
\label{sec:preliminary}

We formulate SparkVLA as a hierarchical VLA with a high-level VLM
planner and a low-level VLA executor.  At a decision boundary~$t$,
defined as a re-observation point at which a new execution decision is made,
the robot receives camera observations $I_t^{(1)},\ldots,I_t^{(C)}$ from
$C$~cameras and proprioceptive state $\state_t\in\mathbb{R}^{d_s}$,
including the end-effector pose and gripper width.  At the beginning of each
subtask, the planner autoregressively generates a natural-language subtask
command~$c_k$, whose
complete inputs are specified below.  Conditioned
on~$c_k$, the current camera observations, and proprioceptive state, the
executor produces an action chunk
$\chunk_t=(\action_t,\ldots,\action_{t+H-1})\in\mathbb{R}^{H\times d_a}$.

We define a unified candidate set $\mathcal{C}=\{0,1,\ldots,H\}$, where
candidate~$0$ is \textbf{Stop} and each candidate $n\geq1$ denotes the
\textbf{action-prefix} consisting of the first $n$ actions in $\chunk_t$.
The selector ranks these candidates from the action chunk and two context
representations, $\alpha_k$ and $m_t$, described next:
\begin{equation}
    \label{eq:selector_overview}
    \selector(\chunk_t,\alpha_k,m_t)\to \mathcal{C}.
\end{equation}
Selecting an action-prefix executes its $n$ actions before re-observing,
whereas selecting Stop completes the current subtask and triggers
autoregressive generation of the next subtask command.
The following subsections describe the closed-loop
architecture, context construction, and candidate scoring, respectively.

\subsection{SparkVLA Architecture}
\label{sec:architecture}

SparkVLA derives two separately parameterized branches from the
pre-trained $\pi_{0.5}$ VLA.  The high-level branch is built from the VLM
backbone of $\pi_{0.5}$ and serves as the autoregressive subtask
generator, whereas the low-level branch uses the complete $\pi_{0.5}$
VLA, including its action expert, to propose action chunks.  The
action-prefix selector is implemented on the high-level planner branch
and is invoked at every decision boundary.

At the beginning of subtask~$k$, the planner autoregressively generates
$c_k$ from the global instruction, the ordered semantic history
$\hist_k=(c_1,\ldots,c_{k-1})$, and the current observations and state.
This history records previously completed subtask commands as an explicit
semantic memory.  Conditioned on $c_k$, the executor proposes $\chunk_t$
at each decision boundary.  If the selector chooses an action-prefix,
its actions are executed before the robot re-observes and reaches the
next decision boundary.  If it chooses Stop, $c_k$ is completed and
appended to the history, $\hist_{k+1}=(\hist_k,c_k)$, after which the
planner generates $c_{k+1}$ from the updated history, current camera
observations, and proprioceptive state.

\subsection{Anchor-Conditioned Context Encoding}
\label{sec:framework}

The selector context is constructed in three stages: extracting a
history-aware subtask anchor~$\alpha_k$, pruning visual tokens under
anchor guidance, and computing the high-level hidden feature~$m_t$.

\paragraph{History-Aware Subtask Anchor.}
At subtask onset, after the planner generates command~$c_k$, we extract
a raw anchor from the penultimate transformer layer.  Let
$h^{(\ell)}_{c_k,\mathrm{last}}$ denote the hidden state at the final
generated token from layer~$\ell$.  Because this representation is
conditioned on the global instruction, semantic subtask history,
subtask-onset observations, proprioceptive state, and the complete
generated command, it encodes both a persistent memory of the onset
environment and a semantic expectation of the subtask goal:
\begin{equation}\label{eq:anchor_raw}
    \hat{\alpha}_k = \mathrm{MLP}_{\mathrm{anchor}}\bigl(
    h^{(\ell_{\mathrm{deep}})}_{c_k,\mathrm{last}}\bigr)
    \in \mathbb{R}^{d}.
\end{equation}
A single raw anchor captures goal semantics and onset-state memory but
lacks awareness of how preceding subtasks have shaped the workspace.  To
propagate cross-subtask context, we maintain an anchor history pool
$\mathcal{A}_k = \{\hat{\alpha}_1,\ldots,\hat{\alpha}_{k-1}\}$ and fuse
the current raw anchor with its history via self-attention:
\begin{equation}\label{eq:anchor_history}
    \alpha_k = \bigl[\mathrm{SelfAttn}\bigl(
    [\hat{\alpha}_1;\ldots;\hat{\alpha}_{k-1};\hat{\alpha}_k]\bigr)
    \bigr]_{\mathrm{last}} \in \mathbb{R}^{d},
\end{equation}
where $[\cdot]_{\mathrm{last}}$ extracts the output at the position of
$\hat{\alpha}_k$.  Through self-attention, $\alpha_k$ inherits
information about which regions have been manipulated in earlier
subtasks---context critical for memory-dependent conditions.  For the
first subtask ($k{=}1$), self-attention reduces to a linear projection.
The fused anchor is computed once per subtask and cached throughout.

\paragraph{Anchor-Guided Visual Pruning.}
The frozen SigLIP~\cite{siglip} encoder produces dense visual tokens
$V_t^{(i)} \in \mathbb{R}^{N \times d_v}$ for each camera~$i$.  We
prune to $K$ task-relevant tokens using
Grid-Sampler~\cite{grid-sample}, a differentiable operator that predicts
$K$ normalized 2D coordinates from a conditioning representation and
retrieves features via bilinear interpolation.  We condition on the
subtask anchor~$\alpha_k$, directing sampling toward subtask-relevant
regions:
\begin{equation}\label{eq:coords}
\begin{aligned}
    \boldsymbol{\xi}_t^{(i)}
        &= \mathrm{MLP}_{\mathrm{coord}}\bigl(
        \mathrm{CrossAttn}(\alpha_k,\, V_t^{(i)})\bigr)
        \in [0,1]^{K\times 2},\\
    \tilde{V}_t^{(i)}
        &= \mathrm{BilinearSample}\bigl(
        V_t^{(i)},\boldsymbol{\xi}_t^{(i)}\bigr)
        \in \mathbb{R}^{K\times d_v}.
\end{aligned}
\end{equation}
Here, $\mathrm{CrossAttn}$ produces an anchor-conditioned visual summary
that $\mathrm{MLP}_{\mathrm{coord}}$ maps to sampling
coordinates~$\boldsymbol{\xi}_t^{(i)}$, reducing the visual sequence
from $N$ to $K$ tokens per camera.

\paragraph{High-Level Hidden Feature Encoding.}
The pruned visual tokens are projected into the backbone width
($\bar{V}_t^{(i)} = \tilde{V}_t^{(i)} W_v \in \mathbb{R}^{K \times d}$),
concatenated with proprioceptive-state and subtask-command embeddings,
and processed through the first $\ell_m$~layers of the high-level VLM
backbone with LoRA~\cite{lora} adapters:
\begin{equation}\label{eq:midfeature}
\begin{aligned}
    X_t &= [\bar{V}_t^{(1)}; \ldots; \bar{V}_t^{(C)};
        \mathrm{MLP}_s(\state_t); \mathrm{Embed}(c_k)],\\
    m_t &= \bigl[\mathrm{Backbone}^{(1:\ell_m)}(X_t)
        \bigr]_{c_k,\mathrm{last}}.
\end{aligned}
\end{equation}
A prefix-attention mask aggregates the multimodal context into
$m_t \in \mathbb{R}^{d}$.  Pruning and the shallow pass are recomputed
at each decision boundary.


\begin{figure*}[t]
\centering
\includegraphics[width=\textwidth]{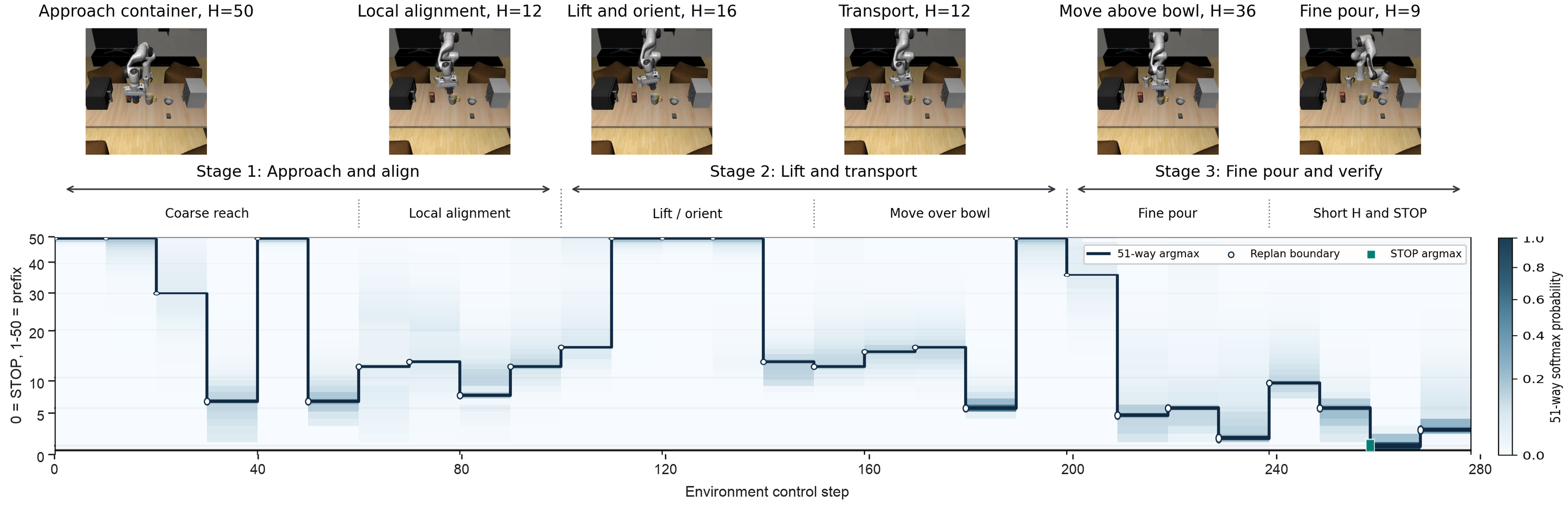}
\caption{Stop-aware action-prefix selection over a representative
pouring episode.  Top: observation frames at decision boundaries with
selected prefix length.  Bottom: the selected candidate (solid line) over
environment steps, with softmax probability shown as a heatmap
(darker = higher probability).  Replan boundaries (circles) mark
subtask transitions.  Long prefixes dominate during coarse reach and
transport, short prefixes appear during fine alignment, and Stop fires at
task completion (green square).}
\label{fig:prefix_distribution}
\end{figure*}

\subsection{Stop-Aware Action-Prefix Selection}
\label{sec:selector}

At each decision boundary the selector receives the action chunk
$\chunk_t$ proposed by the low-level executor, the cached subtask
anchor~$\alpha_k$, and the current high-level hidden feature~$m_t$.
It maps these inputs to scores over the unified candidate set and
selects either Stop or the highest-scoring action-prefix to execute
before re-observing:
\begin{align}
    \mathbf{s}_t &= \selector(\chunk_t, \alpha_k, m_t)
        \in \mathbb{R}^{H+1}, \label{eq:selector_interface}\\
    \hat{n}_t &= \arg\max_{i \in \mathcal{C}}\,s_{t,i}.
        \label{eq:inference}
\end{align}

\paragraph{Unified Candidate Scoring.}
The scoring head is a full-attention transformer in which Stop and
all action-prefix candidates attend to one another.  Each candidate can
therefore evaluate its length against the complete proposed chunk,
including what later actions would accomplish.
The anchor and current high-level hidden feature are projected into
context tokens:
\begin{align}
    e_\alpha &= W_\alpha\,\mathrm{LayerNorm}(\alpha_k),
    \label{eq:ctx_anchor}\\
    e_m &= W_m\,\mathrm{LayerNorm}(m_t),
    \label{eq:ctx_hidden}
\end{align}
where $W_\alpha, W_m \in \mathbb{R}^{d_r \times d}$ project into the
scoring space of dimension~$d_r$.  The Stop candidate uses a learned
embedding $e_0 \in \mathbb{R}^{d_r}$, and subtask and multimodal context
are supplied by $e_\alpha$ and $e_m$ through self-attention.  Each
length-$n$ candidate token is initialized from the terminal action of
its prefix plus a learned length embedding:
\begin{equation}\label{eq:candidate}
    e_n = \mathrm{MLP}_{\action}(\action_{t+n-1}) + p_n,
    \quad n=1,\ldots,H.
\end{equation}
Full self-attention over the complete token set
contextualizes each candidate:
\begin{equation}\label{eq:bitr}
    \mathbf{h} = \mathrm{Transformer}(\mathbf{e}),
    \quad \mathbf{e} = [e_\alpha, e_m, e_0, \ldots, e_H].
\end{equation}
A scoring layer maps each candidate's output $h_i$ to a
scalar $s_{t,i} = \mathrm{MLP}_{\mathrm{score}}(h_i)\in\mathbb{R}$,
$i = 0,\ldots,H$.
The context outputs $h_\alpha$ and $h_m$ provide conditioning
only, and scores are computed from the candidate representations
$h_0,\ldots,h_H$.  Although $e_0$ is fixed, self-attention makes the
resulting Stop representation context-dependent.  Each invocation
requires one anchor-guided pruning step, one shallow backbone pass, and
one compact scoring-head pass.  Anchor extraction is amortized over the
entire subtask.  Figure~\ref{fig:prefix_distribution} visualizes the
resulting selection behavior over a full episode.

\paragraph{Stop-Aware Ranking Objective.}
We optimize the selector with two complementary losses over the same
candidate scores: a pairwise ranking loss for relative ordering, and a
stop-aware auxiliary loss that strengthens the termination decision.
For pairwise ranking, we assign ordinal priorities $\rank_i$ from offline
execution outcomes.  Let $\tau$ denote the subtask completion boundary
and $n^\star = \tau - t$ the remaining actions.  The priority design
encodes three principles:
(1)~after completion ($t \geq \tau$), Stop ranks highest;
(2)~before completion, covering prefixes ($n \geq n^\star$) outrank
non-covering ones, and among covering prefixes the one ending closest to
the boundary is preferred;
(3)~for unsuccessful rollouts where $\tau$ is undefined, all
action-prefixes tie above Stop, encouraging continued exploration
(the episode-level horizon bound guarantees safe termination).
This construction leverages the temporal structure of successful
demonstrations: because expert trajectories reach the subtask boundary
efficiently, the alignment $n^\star = \tau - t$ provides a natural
ordinal supervision signal---a standard inductive bias in offline
preference learning.
During training, we apply uniform random perturbation
$\tau' = \tau + \Delta,\; \Delta \sim \mathcal{U}(-k, k)$ to boundary
annotations, making the ordinal labels robust to annotation noise.
The full priority assignment is provided in the supplementary material.

From the target priorities, we form comparable pairs
$\mathcal{P}_t = \{(i,j): \rank_i \neq \rank_j\}$ with labels
$y_{ij} = \mathbb{1}[\rank_i > \rank_j]$.  The ranking loss is:
\begin{equation}\label{eq:rankloss}
    \mathcal{L}_{\mathrm{rank}}
    = \frac{1}{|\mathcal{P}_t|}
    \sum_{(i,j)\in\mathcal{P}_t}
    \BCE\!\bigl(s_{t,i} - s_{t,j},\; y_{ij}\bigr).
\end{equation}
Since termination events occur only at subtask boundaries, we add
a stop-aware auxiliary to provide stronger gradient signal for the
stop decision:
\begin{align}
    u_t &= s_{t,0} - \log\!\textstyle\sum_{n=1}^{H}\exp(s_{t,n}),
    \label{eq:stop_logit}\\
    \mathcal{L}_{\mathrm{stop}} &=
        \BCE(u_t,\,y_t^{\mathrm{stop}}),
    \label{eq:stop_loss}
\end{align}
where $u_t$ contrasts the Stop score against a log-sum-exp aggregation
of continuation scores, and
$y_t^{\mathrm{stop}} = \mathbb{1}[t \geq \tau]$.  The total loss is
$\mathcal{L} = \mathcal{L}_{\mathrm{rank}}
+ \lambda_{\mathrm{stop}}\,\mathcal{L}_{\mathrm{stop}}$.
Both objectives shape the same scores, so inference uses the unified
rule in Eq.~\eqref{eq:inference};
$\mathcal{L}_{\mathrm{stop}}$ sharpens the stop boundary without
defining a separate inference pathway.
A short confirmation window filters transient score fluctuations before
committing a stop decision.


\section{Experiments}
\label{sec:exp}

\begin{table*}[!t]
\centering
\begin{tabular*}{\textwidth}{@{\extracolsep{\fill}\hspace{8pt}}llccccccc@{\hspace{8pt}}}
\toprule
\textbf{Category} & \textbf{Method} & \textbf{Avg} &
\multicolumn{2}{c}{\textbf{Dynamic}} &
\multicolumn{2}{c}{\textbf{Memory}} &
\textbf{Mix} & \textbf{Ideal} \\
\cmidrule(lr){4-5}\cmidrule(lr){6-7}
& & & Ran. & Obs. & Exp. & Exe. & & \\
\midrule
\multirow{3}{*}{Standalone VLA}
& OpenVLA-OFT & 1.68 & 3.95 & 0.00 & 0.84 & 0.96 & 1.04 & 3.95 \\
& SmolVLA & 2.80 & 5.90 & 4.33 & 0.95 & 1.51 & 0.52 & 5.90 \\
& $\pi_{0.5}$ & 12.29 & 11.84 & 9.09 & 12.61 & 10.58 & 17.71 & 10.53 \\
\midrule
\multirow{5}{*}{Hierarchical VLA}
& HPE & 16.55 & 18.63 & 19.18 & 9.06 & 17.83 & 13.21 & 21.10 \\
& Mem-0 & 20.29 & 18.42 & 18.42 & 25.21 & 18.27 & 19.81 & 19.74 \\
& SparkVLA + OpenVLA-OFT & 24.15 & 18.22 & 20.99 & 28.56 & 26.63 & 27.54 & 18.22 \\
& SparkVLA + SmolVLA & 25.32 & 18.22 & 19.07 & 30.68 & 30.26 & 27.54 & 19.88 \\
& \textbf{SparkVLA (ours)} & \textbf{47.12} & \textbf{33.13} & \textbf{43.87} & \textbf{60.30} & \textbf{43.58} & \textbf{48.52} & \textbf{46.38} \\
\bottomrule
\end{tabular*}
\caption{Main comparison on RoboCerebra (SR~\%).  \emph{Avg}
is the micro-average across all conditions.}
\label{tab:main_results}
\end{table*}

\paragraph{Training.}
Both the high-level planner and low-level executor are independently
fine-tuned from pre-trained $\pi_{0.5}$ checkpoints on target
manipulation data.  The high-level branch is then trained in two
phases: we first supervise the planner backbone for subtask command
prediction, then freeze both base branches and train the selector
modules to jointly optimize stop and dynamic chunk selection via the
unified ranking objective.  Further implementation details are
provided in the supplementary material.

\subsection{Simulation Experiments}
\label{sec:sim}

\paragraph{Benchmarks.}
Our primary benchmark is RoboCerebra, a long-horizon manipulation
benchmark with 1{,}000 annotated trajectories spanning 100 task
variants (avg.\ 9.1 subtasks, 2{,}972 steps).  Six conditions test
complementary capabilities: \emph{Dynamic} (randomized layouts /
changing observations), \emph{Memory} (subtask-history recall),
\emph{Mix}, and a static \emph{Ideal} control.  We use 60 held-out
cases with ten rollouts each (600 episodes total).
We additionally evaluate on LIBERO~\cite{libero}(Spatial, Object, Goal, Long; 40
tasks, 100 episodes each) with subtask annotations from
LaRA-VLA~\cite{LaRA-VLA}.

\paragraph{Metrics.}
Following the official RoboCerebra protocol, we report \emph{Success
Rate}~(SR) as the primary metric, defined as the fraction of evaluation
episodes in which the robot completes the full task.  \emph{Avg} denotes
the micro-average SR across all six conditions.  We additionally report
\emph{Stop Acc.}, the agreement between the selector's stop decisions
and annotated subtask boundaries, measuring how precisely the system
identifies subtask completion.

\paragraph{Baselines.}
We compare two paradigms.
(i)~\textbf{Standalone VLA}: OpenVLA-OFT~\cite{openvla_oft}, SmolVLA~\cite{smolvla}, and $\pi_{0.5}$~\cite{pi05},
each receiving a global task prompt without subtask decomposition.
(ii)~\textbf{Hierarchical VLA}: HPE~\cite{robocerebra}, the
hierarchical planning-execution framework whose results are reported in
the original RoboCerebra paper; Mem-0~\cite{rmbench}, a recent
open-source hierarchical VLA that we reproduce on RoboCerebra.
Two additional SparkVLA variants swap the low-level executor
(OpenVLA-OFT, SmolVLA) to test generalization.
We focus on baselines that are publicly accessible and technically
reproducible on RoboCerebra.

\begin{table}[!t]
\centering
\begin{tabular}{@{}lccccc@{}}
\toprule
\textbf{Method} & \textbf{Spa.} & \textbf{Obj.} & \textbf{Goal} &
\textbf{Long} & \textbf{Avg.} \\
\midrule
Diffusion Policy & 78.3 & 92.5 & 68.3 & 50.5 & 72.4 \\
OpenVLA & 84.7 & 88.4 & 79.2 & 53.7 & 76.5 \\
SmolVLA & 93.0 & 94.0 & 91.0 & 77.0 & 88.8 \\
OpenVLA-OFT & 97.6 & 98.4 & 97.9 & 94.5 & 97.1 \\
$\pi_{0.5}$ & \textbf{99.4} & \textbf{98.2} & 97.8 & 92.4 & 96.9 \\
SparkVLA (ours) & 99.2 & 97.6 & \textbf{99.2} & \textbf{98.0} & \textbf{98.5} \\
\bottomrule
\end{tabular}
\caption{Task-level success rate (\%) on LIBERO.}
\label{tab:libero_results}
\end{table}

\paragraph{Results.}
Table~\ref{tab:main_results} summarizes results on RoboCerebra.
SparkVLA achieves 47.12\% average SR, outperforming standalone
$\pi_{0.5}$ by 34.83\% and the official RoboCerebra baseline
(HPE) by 30.57\%.  SparkVLA also surpasses the independently
reproduced Mem-0 by 26.83\%, confirming its advantage over
recent open-source hierarchical methods.  The largest gains appear on \emph{Dynamic Obs.}\
(+34.78\%) and \emph{Memory Exp.}\ (+47.69\%): adaptive shorter
prefixes enable more frequent re-observation under changing scenes,
while the history-enriched anchor prevents revisiting completed targets.
Figure~\ref{fig:prefix_distribution} visualizes this behavior.
Gains are consistent across executors (+22.47 and +22.52\%) over
standalone OpenVLA-OFT and SmolVLA), confirming that SparkVLA's
high-level system generalizes to different low-level executors.  We isolate the
contribution of each design choice in the ablation below.

On LIBERO (Table~\ref{tab:libero_results}), SparkVLA reaches 98.5\%
average success.  Most suites are near ceiling for strong baselines;
the clearest separation is on LIBERO-Long (5.6\%), where multi-step
sequences benefit most from subtask decomposition.

\begin{figure*}[!t]
\centering
\includegraphics[width=0.90\textwidth]{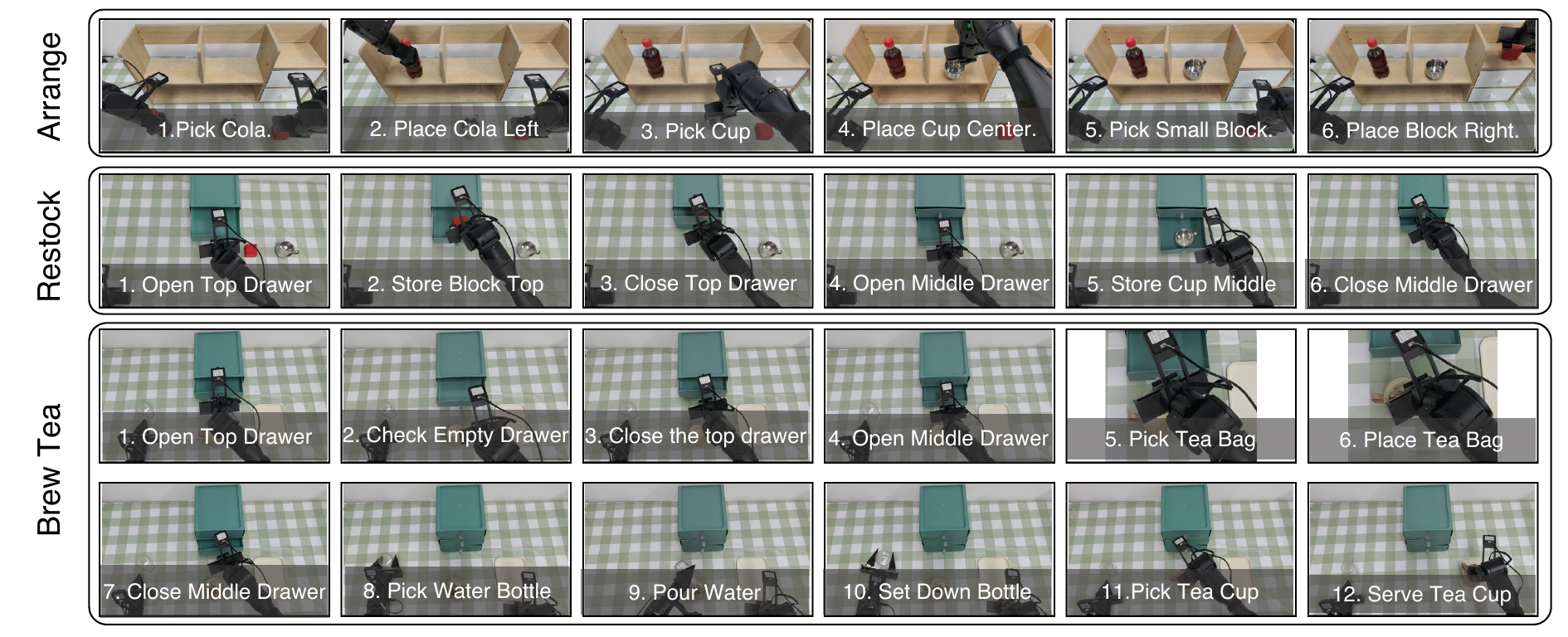}
\caption{Real-world task rollouts.  Top to bottom: Shelf Arrangement
(6 subtasks), Drawer Restocking (6 subtasks), Tea Preparation
(12 subtasks).}
\label{fig:real_world}
\end{figure*}

\subsection{Ablation Studies}
\label{sec:ablation}

We ablate chunk-selection method, unified stop integration,
context encoding with selector placement, and inference scheduling.
All variants use the same RoboCerebra protocol.

\begin{table}[t]
\centering
\begin{tabular*}{0.97\columnwidth}{@{\extracolsep{\fill}}lcc@{}}
\toprule
\textbf{Configuration} &
\textbf{SR} & \textbf{Stop Acc.} \\
\midrule
\multicolumn{3}{@{}l}{\textit{(a) Chunk selection (independent stop head)}} \\
Fixed              & 34.26 & 79.52 \\
RLPD               & 37.62 & 80.05 \\
AQC                & 40.92 & 80.43 \\
Pairwise ranking   & 42.53 & 80.37 \\
\midrule
\multicolumn{3}{@{}l}{\textit{(b) Stop integration (pairwise ranking fixed)}} \\
Unified candidate set (ours) & \textbf{47.12} & \textbf{96.65} \\
\bottomrule
\end{tabular*}
\caption{Ablation of chunk selection and stop integration.
Block~(a) varies only the selection method while keeping an independent stop head;
block~(b) unifies stop into the same candidate set, yielding gains in \emph{both} SR and Stop Acc.}
\label{tab:ablation_chunk}
\end{table}

\paragraph{Chunk selection and stop integration.}
Table~\ref{tab:ablation_chunk} isolates two design axes.
Block~(a) fixes an independent stop head and varies only the
chunk-selection method: upgrading from a fixed horizon to
RLPD~\cite{rlpd} and AQC~\cite{aqc} progressively raises SR from
34.26\% to 40.92\%; our pairwise ranking further yields 42.53\%.
Notably, Stop Acc.\ remains near 80\% throughout, confirming that the
independent head cannot benefit when the two decisions are decoupled.
Block~(b) keeps pairwise ranking and unifies \textsc{Stop} into the same
candidate set: Stop Acc.\ jumps from 80.37\% to 96.65\% (+16.28\%,
precision 94.3\%, recall 98.1\%) while SR rises from 42.53\% to 47.12\%
, confirming bidirectional benefit: stop decisions gain from
action-prefix context, and prefix selection gains from an accurate
termination signal.

\paragraph{Context encoding and selector placement.}
Table~\ref{tab:ablation_context} isolates visual-token processing and
backbone choice.  Progressive improvements from no pruning (40.03\%) to
spatial pruning (43.46\%), raw anchor (45.18\%), and the full
history-aware anchor (47.12\%) confirm that each stage contributes; the
history-aware gain (1.94\%) comes from resolving spatial ambiguity in
memory-dependent conditions.  For selector placement, the planner
backbone adds 9.84\% over the executor backbone, because the
planner encodes richer task-progress and subtask-history context
that the executor which trained only for action generation does not retain.

\begin{table}[t]
\centering
\begin{tabular}{@{\hspace{6pt}}lcc@{\hspace{6pt}}}
\toprule
\textbf{Variant} & \textbf{SR} & \textbf{Stop Acc.} \\
\midrule
\multicolumn{3}{@{\hspace{6pt}}l}{\textit{Visual token processing}} \\
\midrule
No pruning & 40.03 & 88.82 \\
Spatial pruning only & 43.46 & 92.16 \\
Raw anchor only & 45.18 & 94.03 \\
History-aware anchor (ours) & \textbf{47.12} & \textbf{96.65} \\
\midrule
\multicolumn{3}{@{\hspace{6pt}}l}{\textit{Selector backbone}} \\
\midrule
Low-level executor & 37.28 & 92.60 \\
High-level planner (ours) & \textbf{47.12} & \textbf{96.65} \\
\bottomrule
\end{tabular}
\caption{Ablation of anchor design and selector placement.}
\label{tab:ablation_context}
\end{table}

\paragraph{Inference scheduling.}
Two runtime optimizations improve efficiency: caching the subtask
command at subtask onset provides a $2.91\times$ speedup with no
accuracy loss, and invoking the selector only at chunk boundaries (vs.\
every step) yields $2.10\times$ speed while improving Stop Acc.\ from
93.10\% to 96.65\% (+3.55~pp), since stationary inter-chunk frames are
free of motion blur.  With both optimizations, SparkVLA runs at
4.14\,Hz (41.4 action steps/s) vs.\ 4.68\,Hz for standalone
$\pi_{0.5}$, retaining 88\% throughput.

\subsection{Real-World Experiments}
\label{sec:realworld}

To validate whether the adaptive prefix selection and stop-aware
termination transfer to physical hardware, we deploy SparkVLA on an
AirbotPlay dual-arm platform with two wrist and one head RGB cameras across three tasks
(Figure~\ref{fig:real_world}):
(1)~\emph{Shelf Arrangement} (6 subtasks), repeated pick-and-place
requiring persistent spatial binding;
(2)~\emph{Drawer Restocking} (6 subtasks), articulated-object
interaction with drawer-state tracking; and
(3)~\emph{Tea Preparation} (12 subtasks), combining exploration,
observation-based confirmation, constrained pouring, and long-horizon
chaining.
We collect 50 demonstrations per task and evaluate over 50 rollouts
with matched initial states.  Success requires completing all subtasks
and satisfying final-state conditions.

\begin{table}[!ht]
\centering
\begin{tabular}{@{}lcccc@{}}
\toprule
\textbf{Method} & \textbf{Arr.} & \textbf{Restock} &
\textbf{Brew Tea} & \textbf{Avg.} \\
\midrule
OpenVLA-OFT & 44.0 & 36.0 & 10.0 & 30.0 \\
$\pi_{0.5}$ & 64.0 & 54.0 & 26.0 & 48.0 \\
SparkVLA (ours) & \textbf{78.0} & \textbf{74.0} & \textbf{56.0} & \textbf{69.3} \\
\bottomrule
\end{tabular}
\caption{Real-world success rate (\%) on three multi-step tasks.}
\label{tab:real_world_results}
\end{table}

SparkVLA achieves 69.3\% average success, outperforming $\pi_{0.5}$ by
+21.3\%(Table~\ref{tab:real_world_results}).  The largest margin
appears on Tea Preparation (+30.0\%), where the 12-subtask horizon
magnifies the benefit of explicit subtask memory and stop-aware
termination.  Across all tasks, adaptive prefixes shorten execution near
contact events and extend it during free-space transport---the same
behavioral pattern observed in simulation
(Figure~\ref{fig:prefix_distribution}), confirming that the learned
policy transfers qualitatively to physical dynamics.

\section{Conclusion}

We have shown that subtask termination and execution-horizon selection
in hierarchical VLA systems are mutually dependent decisions that
existing architectures evaluate in isolation, creating a structural
information asymmetry at the planning-execution interface. We have proposed SparkVLA, a novel hierarchical VLA framework
resolving this by placing both decisions in a single unified candidate
set and selecting the highest-scoring option, which is trained entirely
from offline ordinal preferences.  Experiments on RoboCerebra, LIBERO, and real-robot tasks
demonstrate substantial improvements over both standalone and
hierarchical baselines, with consistent executor-agnostic transfer
confirming that the gains derive from the unified interface design
rather than a particular model pairing.
The current evaluation focuses on tabletop manipulation with fixed
object categories and offline training; extending SparkVLA to
open-world settings, online adaptation, and a broader range of
planner--executor combinations remains a promising future direction.

\bibliography{aaai2027}


\clearpage
\section*{Supplementary Material}

This document supplements the main paper with the details needed to
interpret and reproduce its central claims.It covers the complete
ordinal supervision, paper-aligned implementation settings, closed-loop
inference, common evaluation controls, physical-system protocol, and
representative simulation scenes.  Definitions and results already
given in the main paper are not repeated.

\section{Complete Ordinal Supervision}

\subsection{Priority Assignment}

The main paper gives the principles used to supervise the unified
candidate set.  Here we provide the complete assignment, including
trajectory-end masking and failure rollouts.  Let $T$ be the number of
recorded actions in a training trajectory, and define candidate
validity by
\begin{equation}
    v_0=1, \qquad
    v_n=\mathbb{1}[t+n-1<T], \quad n=1,\ldots,H.
\end{equation}
For compactness, write $\mathrm{succ}$ and $\mathrm{fail}$ for
successful and unsuccessful trajectories.  For a successful
trajectory, we perturb its annotated completion
boundary during training as $\tau'=\tau+\Delta$, where
$k=2$ and $\Delta$ is sampled uniformly from
$\{-2,-1,0,1,2\}$, and set
$n^\star=\tau'-t$.  The Stop priority is
\begin{equation}
\rho_0=
\begin{cases}
H+1, & \mathrm{succ},\ t\geq\tau',\\
0,   & \text{otherwise}.
\end{cases}
\label{eq:supp_stop_priority}
\end{equation}
For each valid action-prefix candidate $n\in\{1,\ldots,H\}$,
\begin{equation}
\rho_n=
\begin{cases}
H-n+1,
  & \mathrm{succ},\ t\geq\tau',\\
n,
  & \mathrm{succ},\ t<\tau',\ n^\star>H,\\
n,
  & \mathrm{succ},\ t<\tau',\ n<n^\star\leq H,\\
2H-(n-n^\star),
  & \mathrm{succ},\ t<\tau',\ n^\star\leq n\leq H,\\
1,
  & \mathrm{fail}.
\end{cases}
\label{eq:supp_prefix_priority}
\end{equation}
Thus, after completion, Stop outranks every action-prefix.  Before a
reachable boundary, every covering prefix outranks every non-covering
prefix, and lower overshoot is preferred among covering prefixes.  If
the boundary lies beyond the proposed chunk ($n^\star>H$), longer
prefixes receive higher priority.  In an unsuccessful rollout, all
action-prefixes tie above Stop; such samples supervise continuation but
do not impose an arbitrary preferred execution length.

Only valid candidates with unequal priorities form a training pair:
\begin{equation}
\mathcal{P}_t = \{(i,j)\mid 0\leq i<j\leq H,\;
v_i v_j=1,\;\rho_i\neq\rho_j\}.
\label{eq:supp_pairs}
\end{equation}
The strict upper-triangular construction in
Eq.~\eqref{eq:supp_pairs} includes each unordered comparison once.
Consequently, tied action-prefixes in unsuccessful rollouts do not
enter the pairwise loss, whereas their Stop-versus-prefix comparisons
remain active.  Unsuccessful-rollout ranking samples are assigned a
weight of $0.1$.  Algorithm~\ref{alg:supp_priority} gives the complete
procedural construction used by the released preprocessing code.

\begin{algorithm}[tb]
\caption{Ordinal supervision for unified candidates}
\label{alg:supp_priority}
\textbf{Input}: decision step $t$, trajectory metadata
$(T,q,\tau)$ (with $\tau$ defined only if $q=1$), horizon $H$\\
\textbf{Output}: priorities $\boldsymbol{\rho}$, validity
$\mathbf{v}$, Stop label $y^{\mathrm{stop}}$, weights
$w^{\mathrm{rank}},w^{\mathrm{stop}}$, pair set $\mathcal{P}_t$
\begin{algorithmic}[1]
\STATE Initialize $\boldsymbol{\rho}\gets\mathbf{0}$,
$\mathbf{v}\gets\mathbf{0}$, and $v_0\gets1$
\FOR{$n=1,\ldots,H$}
    \STATE $v_n\gets\mathbb{1}[t+n-1<T]$
\ENDFOR
\IF{$q=0$}
    \STATE $\rho_0\gets0$ and $\rho_n\gets1$ for each valid $n\geq1$
    \STATE $y^{\mathrm{stop}}\gets0$,
    $w^{\mathrm{rank}}\gets0.1$, $w^{\mathrm{stop}}\gets1.5$
\ELSE
    \STATE Sample $\Delta$ uniformly from $\{-2,-1,0,1,2\}$
    \STATE $\tau'\gets\tau+\Delta$ and $n^\star\gets\tau'-t$
    \STATE $w^{\mathrm{rank}}\gets1$
    \IF{$t\geq\tau'$}
        \STATE $\rho_0\gets H+1$ and
        $\rho_n\gets H-n+1$ for each valid $n\geq1$
        \STATE $y^{\mathrm{stop}}\gets1$ and
        $w^{\mathrm{stop}}\gets3$
    \ELSE
        \STATE $\rho_0\gets0$ and $y^{\mathrm{stop}}\gets0$
        \STATE $w^{\mathrm{stop}}\gets
        3$ if $n^\star\leq80$, and $1$ otherwise
        \FOR{each valid $n=1,\ldots,H$}
            \IF{$n^\star>H$ or $n<n^\star$}
                \STATE $\rho_n\gets n$
            \ELSE
                \STATE $\rho_n\gets2H-(n-n^\star)$
            \ENDIF
        \ENDFOR
    \ENDIF
\ENDIF
\STATE $\mathcal{P}_t\gets
\{(i,j)\mid0\leq i<j\leq H,\ v_iv_j=1,\ \rho_i\neq\rho_j\}$
\STATE \textbf{return}
$\boldsymbol{\rho},\mathbf{v},y^{\mathrm{stop}},
w^{\mathrm{rank}},w^{\mathrm{stop}},\mathcal{P}_t$
\end{algorithmic}
\end{algorithm}

\subsection{Stop-Aware Optimization}

The training objective contains the two terms stated in the main paper:
the pairwise relative-ranking loss and the stop-aware binary loss, both
applied to the same $H+1$ candidate scores.  To counter the sparsity of
completion boundaries without changing the inference rule, Stop-positive samples
and near-boundary Stop-negative samples receive weight $3.0$; negative
samples from unsuccessful rollouts receive weight $1.5$; all remaining
samples use unit weight.  We set $\lambda_{\mathrm{stop}}=1$ in
$\mathcal{L}=\mathcal{L}_{\mathrm{rank}}+
\lambda_{\mathrm{stop}}\mathcal{L}_{\mathrm{stop}}$.  The pairwise
term includes every valid Stop--action-prefix comparison with unequal
priority and therefore directly supervises the score ordering used at
inference.  The stop-aware term adds a boundary-focused aggregate
contrast over the same scores.  At inference, the selector applies the
unified comparison rule from the main paper to this score vector, so the
auxiliary term strengthens boundary supervision while preserving a
single decision pathway.  Algorithm~\ref{alg:supp_optimization}
expands the minibatch computation.  In particular, it does not introduce
an additional prediction head or optimization term.

\begin{algorithm}[tb]
\caption{Two-loss optimization of the unified selector}
\label{alg:supp_optimization}
\textbf{Input}: minibatch of selector inputs and
Algorithm~\ref{alg:supp_priority} targets\\
\textbf{Parameter}: $\lambda_{\mathrm{stop}}=1$\\
\textbf{Output}: $\mathcal{L}$
\begin{algorithmic}[1]
\STATE Initialize $R,Z_R,S,Z_S\gets0$
\FOR{each sample $b=1,\ldots,B$}
    \STATE $\mathbf{s}_b\gets
    f_{\phi,\theta}^{\mathrm{sel}}(A_t,\alpha_k,m_t)$
    \STATE $\mathcal{P}_b\gets
    \{(i,j)\mid 0\leq i<j\leq H,\ v_iv_j=1,\ \rho_i\neq\rho_j\}$
    \FOR{each $(i,j)\in\mathcal{P}_b$}
        \STATE $y_{ij}\gets\mathbb{1}[\rho_i>\rho_j]$
        \STATE $R\gets R+w_b^{\mathrm{rank}}\,
        \BCE(s_{b,i}-s_{b,j},y_{ij})$
        \STATE $Z_R\gets Z_R+w_b^{\mathrm{rank}}$
    \ENDFOR
    \STATE $u_b\gets s_{b,0}-
    \log\sum_{n=1}^{H}\exp(s_{b,n})$
    \STATE $S\gets S+w_b^{\mathrm{stop}}\,
    \BCE(u_b,y_b^{\mathrm{stop}})$
    \STATE $Z_S\gets Z_S+w_b^{\mathrm{stop}}$
\ENDFOR
\STATE $\mathcal{L}_{\mathrm{rank}}\gets R/\max(Z_R,1)$
\STATE $\mathcal{L}_{\mathrm{stop}}\gets S/\max(Z_S,1)$
\STATE $\mathcal{L}\gets\mathcal{L}_{\mathrm{rank}}+
\lambda_{\mathrm{stop}}\mathcal{L}_{\mathrm{stop}}$
\STATE Update the LoRA, context-encoding, and scoring parameters
\STATE \textbf{return} $\mathcal{L}$
\end{algorithmic}
\end{algorithm}

\section{Simulation Training Details}

\subsection{Model Configuration}

Table~\ref{tab:supp_model_config} lists the selector-specific settings
that are omitted from the main paper for space.  The SigLIP visual
encoder and the base parameters of both $\pi_{0.5}$ branches are
frozen during selector training.  Trainable parameters comprise the
high-level LoRA adapters, anchor projection and history attention,
coordinate-prediction module, context projections, and unified scoring
head.

\begin{table}[t]
\centering
\begin{tabular}{@{}p{0.57\columnwidth}p{0.34\columnwidth}@{}}
\toprule
\textbf{Component} & \textbf{Setting} \\
\midrule
Action-chunk horizon $H$ & 50 \\
Visual tokens per camera $K$ & 32 \\
Input image resolution & $224\times224$ \\
Anchor width $d$ & 2,048 \\
High-level depth $\ell_m$ & 8 layers \\
Anchor source $\ell_{\mathrm{deep}}$ & penultimate layer \\
Scoring width $d_r$ & 1,024 \\
Scoring transformer & 2 layers, 8 heads \\
Scoring feed-forward width & 4,096 \\
Scoring dropout & 0 \\
Coordinate context & 4 queries, width 512 \\
LoRA rank / scaling & 16 / 16 \\
LoRA layers & high-level layers 1--8 \\
State normalization & quantile \\
\bottomrule
\end{tabular}
\caption{Selector and context-encoding configuration.}
\label{tab:supp_model_config}
\end{table}

\subsection{Training Data Construction}

Selector training combines 6,660 demonstration records and 5,630
policy-rollout records, from which 2,034,310 decision samples are
constructed.  The mixed sampler draws 30\% demonstration samples and
70\% rollout samples.  For a successful demonstration segment,
$\tau$ is the annotated end of the completed subtask; for a successful
policy rollout, it is the recorded completion step.  Failed rollouts
have no completion boundary and use the last case of
Eq.~\eqref{eq:supp_prefix_priority}.  We retain at most 80 frames after
a successful boundary and cap failed rollouts at 300 frames.  These
failure samples expose the selector to off-demonstration states while
avoiding false Stop supervision.

\subsection{Optimization and Compute}

The planner and executor are initialized independently from the
pre-trained $\pi_{0.5}$ VLA and optimized for subtask-command generation
and action-chunk prediction, respectively.  The autoregressive
subtask-generation phase runs for 2,000 optimizer steps.  Starting from
that planner, selector training runs
for 20,000 optimizer steps in bfloat16 with gradient checkpointing.
We use AdamW, a global batch size of 256, weight decay $0.01$, 1,000
linear warmup steps, gradient-norm clipping at $1.0$, and random seed
7.  Learning rates are $3\times10^{-5}$ for LoRA parameters,
$3\times10^{-4}$ for the scoring and projection heads, and
$1\times10^{-4}$ for the visual coordinate module.  Checkpoints are
saved every 1,000 optimizer steps.

The reported selector run uses four NVIDIA A100 GPUs with 80\,GB memory
each and an Intel Xeon Platinum 8470Q host under Ubuntu~24.04.  The
software environment uses Python~3.11, PyTorch~2.1.2, and CUDA~11.8;
the effective global batch is distributed with NCCL.  The complete
package-level dependency versions are recorded in the Code and Data
Supplement.

\section{Closed-Loop Inference Details}

\subsection{Execution Procedure}

The following procedure makes explicit when each high- and low-level
operation is invoked.  It is a procedural restatement of the model
interface, not an additional inference module.

\begin{enumerate}
    \item Initialize the ordered semantic history as
    $\mathcal{H}_1=()$ and acquire the current camera observations and
    proprioceptive state.
    \item At the onset of subtask $k$, autoregressively generate
    command $c_k$ from the global instruction, $\mathcal{H}_k$, and
    the current robot observation.  Extract and cache the
    history-aware anchor $\alpha_k$.
    \item At decision boundary $t$, query the low-level executor for
    action chunk $A_t$ and recompute the anchor-conditioned high-level
    hidden feature $m_t$.
    \item Score $[\textsc{Stop},1,\ldots,H]$ jointly.  If an
    action-prefix $n$ is selected, execute its first $n$ actions,
    re-observe, and return to Step~3.
    \item When a Stop decision is committed, append $c_k$ to the
    history, set $\mathcal{H}_{k+1}=(\mathcal{H}_k,c_k)$, and
    return to Step~2 for the next subtask, unless the configured global
    termination condition ends the episode.
\end{enumerate}

The selector is invoked only at decision boundaries; neither the
planner nor selector interrupts an action-prefix between two such
boundaries.  This scheduling preserves a closed-loop observation after
every selected prefix while amortizing high-level computation.
Algorithm~\ref{alg:supp_inference} summarizes the full execution loop
and makes explicit that semantic and anchor histories are updated only
after Stop is selected.

\begin{algorithm}[tb]
\caption{SparkVLA closed-loop inference}
\label{alg:supp_inference}
\textbf{Input}: global instruction $g$, planner, executor, unified
selector, horizon $H$\\
\textbf{Output}: executed task trajectory
\begin{algorithmic}[1]
\STATE Initialize semantic history $\mathcal{H}_1\gets()$,
anchor history $\mathcal{A}_1\gets()$, and $k\gets1$
\STATE Acquire current observations $\{I_t^{(i)}\}_{i=1}^{C}$ and
proprioceptive state $s_t$
\WHILE{the global termination criterion is not satisfied}
    \STATE Autoregressively generate
    $c_k\gets f_\phi^{\mathrm{plan}}
    (g,\mathcal{H}_k,\{I_t^{(i)}\},s_t)$
    \STATE $\hat{\alpha}_k\gets\mathrm{MLP}_{\mathrm{anchor}}
    (h^{(\ell_{\mathrm{deep}})}_{c_k,\mathrm{last}})$
    \STATE $\alpha_k\gets
    [\mathrm{SelfAttn}(\mathcal{A}_k,\hat{\alpha}_k)]_{\mathrm{last}}$
    and cache $\alpha_k$
    \WHILE{the current subtask is active}
        \STATE Propose action chunk
        $A_t\gets\pi_{\mathrm{L}}(c_k,\{I_t^{(i)}\},s_t)$
        \STATE Compute anchor-guided visual tokens and shallow
        high-level feature $m_t$
        \STATE $\mathbf{s}_t\gets
        f_{\phi,\theta}^{\mathrm{sel}}(A_t,\alpha_k,m_t)$
        \STATE Mask any action-prefix candidate not provided as valid
        by the executor
        \STATE $\hat{n}_t\gets
        \arg\max_{i\in\{0,\ldots,H\}}s_{t,i}$
        \IF{$\hat{n}_t=0$}
            \STATE $\mathcal{H}_{k+1}\gets(\mathcal{H}_k,c_k)$ and
            $\mathcal{A}_{k+1}\gets
            (\mathcal{A}_k,\hat{\alpha}_k)$
            \STATE $k\gets k+1$ and exit the inner loop
        \ELSE
            \STATE Execute the first $\hat{n}_t$ actions of $A_t$
            \STATE Re-observe $\{I_t^{(i)}\}_{i=1}^{C}$ and $s_t$ at
            the next decision boundary
        \ENDIF
    \ENDWHILE
\ENDWHILE
\STATE \textbf{return} the executed trajectory
\end{algorithmic}
\end{algorithm}

\section{Evaluation Controls}

For each benchmark, all reproduced methods receive the same task
instructions and observation streams and use common action interfaces,
held-out evaluation cases, rollout budget, episode horizon, reset
distribution, and success criteria.  Externally reported results are
included only when they follow the official benchmark protocol and are
identified as such in the main paper.  A common metric definition is
applied to the resulting episode outcomes.  Within each ablation block,
only the named design factor is
changed; the remaining data, training, checkpoint-selection, and
evaluation settings are retained.  These controls make the main
comparison an evaluation of complete systems while allowing the
ablation blocks to isolate individual SparkVLA design choices.

Locally reproduced baselines follow a common task-specific fine-tuning
protocol with matched data volume, training compute, and
hyperparameter-selection budgets.  Planner access and model inputs are
kept consistent within controlled comparisons and otherwise follow the
corresponding method configuration.  Model selection is performed
independently of the evaluation episodes.

Training uses seed~7, while evaluation uses the distinct base seed
2027.  A deterministic per-episode seed is derived from the benchmark,
condition, case identifier, and rollout index, and is applied to Python,
NumPy, PyTorch, and all visible CUDA devices.  For physical trials, the
seed controls trial ordering; task-specific reset ranges define the
initial states.  The derivation function and machine-readable seed
configuration are included in the Code and Data Supplement.

\section{RoboCerebra Task Visualizations}

\begin{figure*}[t]
\centering
\includegraphics[width=\textwidth]{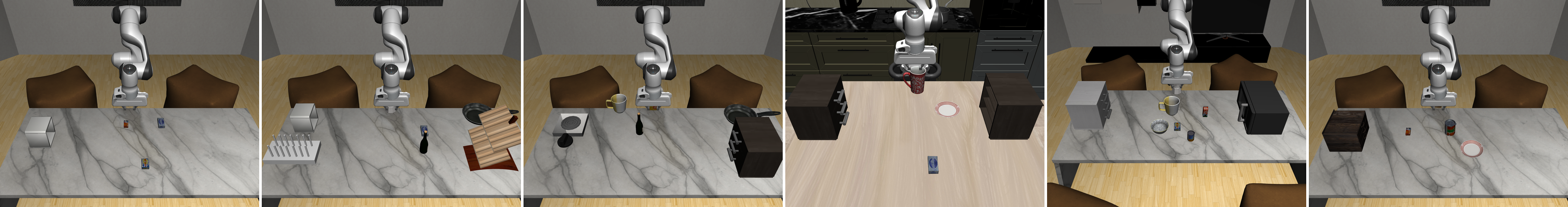}
\includegraphics[width=\textwidth]{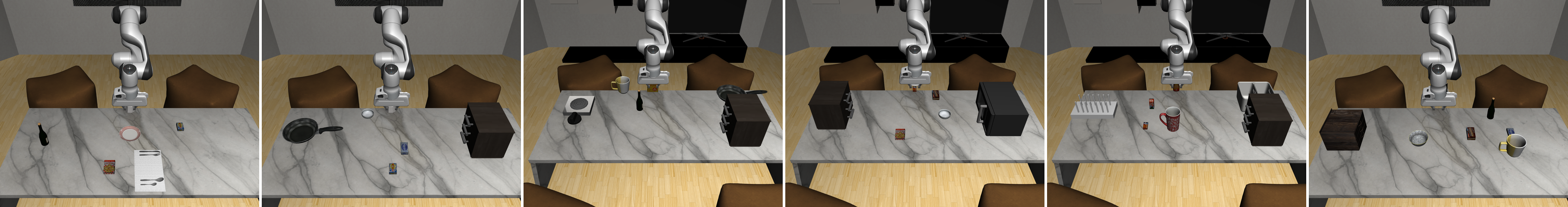}
\includegraphics[width=\textwidth]{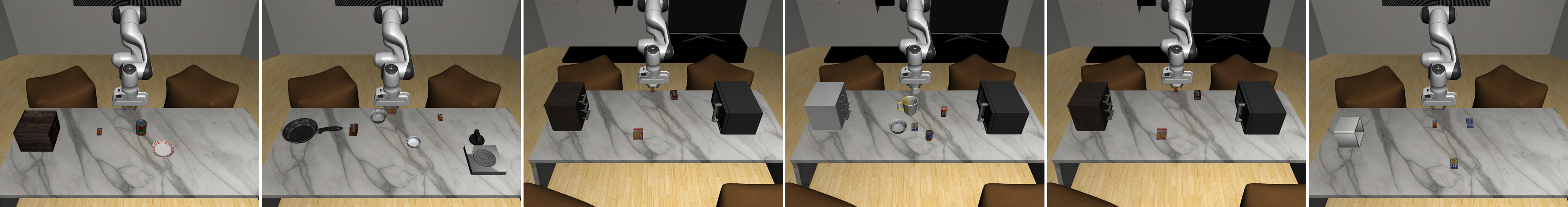}
\includegraphics[width=\textwidth]{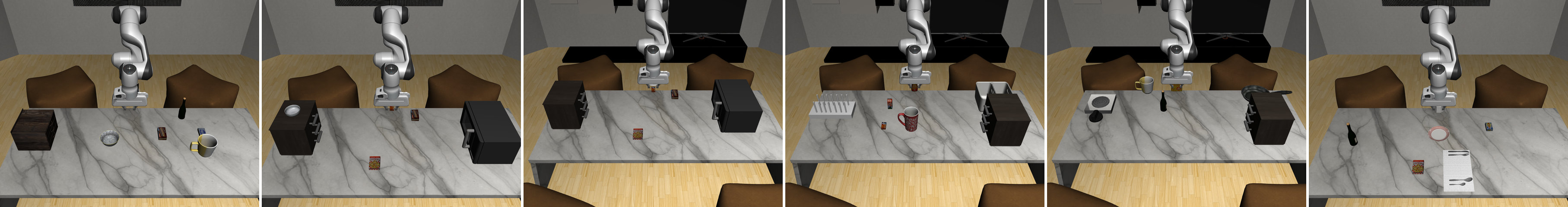}
\includegraphics[width=\textwidth]{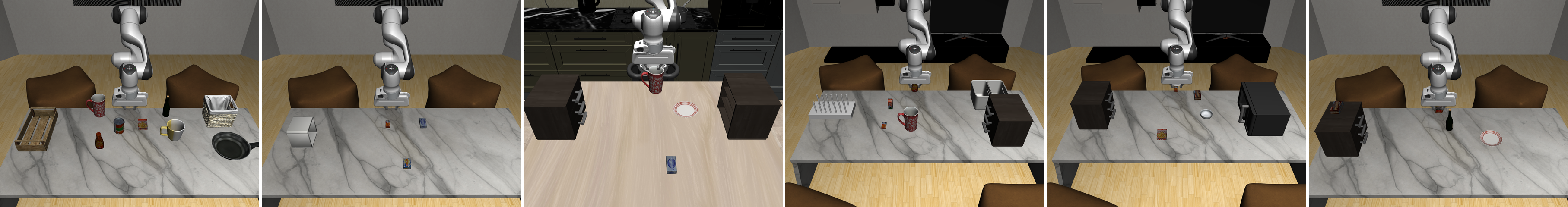}
\caption{Representative RoboCerebra simulation scenes.  From left to
right, the columns show Dynamic Randomization, Dynamic Observation,
Memory Exploration, Memory Execution, Mix, and Ideal; rows contain
distinct held-out task instances.}
\label{fig:supp_robocerebra_tasks}
\end{figure*}

Figure~\ref{fig:supp_robocerebra_tasks} shows representative initial
scenes from the six RoboCerebra conditions used in the main paper.  The
columns preserve the table order in the main paper; the five rows show
different held-out task instances and layouts.  The gallery illustrates
the changes in object placement, distractors, containers, and scene
configuration covered by the benchmark without repeating its metric or
split definitions.

\section{Real-World Experimental Details}

\subsection{Inference Platform and Control Stack}

The real-world evaluation uses a fixed tabletop manipulation platform
with RGB observation streams and proprioceptive feedback.  A
common local deployment stack provides perception, model inference, and
robot control for every compared method.  Thus, sensing and control
conditions remain fixed while only the evaluated policy changes.  Action
commands are streamed at approximately 20\,Hz, corresponding to 50\,ms
per action step.  Model-side planning and selection support approximately
4.14 boundary updates per second, as reported in the main paper; the
realized selector invocation rate is lower and varies with the selected
action-prefix length because the selector runs only at decision
boundaries.  All compared methods use the same action-command rate.

\subsection{Physical Task Definitions}

The physical evaluation contains three ordered long-horizon tasks.
\emph{Shelf Arrangement} contains six subtasks: pick up the cola and
place it in the left shelf compartment; pick up the cup and place it in
the center compartment; then pick up the small block and place it in the
right compartment.  \emph{Drawer Restocking} contains six subtasks:
open the top drawer, store the block, close the top drawer, open the
middle drawer, store the cup, and close the middle drawer.
\emph{Tea Preparation} contains twelve subtasks: open the top drawer,
verify that it is empty, close it, open the middle drawer, retrieve the
tea bag, place it in the cup, close the middle drawer, pick up the water
bottle, pour the bottle contents into the cup, set down the bottle, pick
up the prepared cup, and place it on the serving tray.  These ordered sequences
define the subtask histories and completion events used in data
segmentation and evaluation.  The task-level language retains the
semantic terms \emph{water bottle} and \emph{pour water}.  In the
physical setup, however, the bottle contains white plastic beads, which
serve as a safe and repeatable dry surrogate for water.  The substitution
preserves the manipulation sequence relevant to this task: grasping and
transporting the bottle, aligning it above the cup, controlling its
orientation during transfer, returning it upright, and placing it back
on the table.  The same bottle contents and task setup are used for all
compared methods.

\subsection{Data Collection and Reset Protocol}

Demonstrations are collected with the global instructions and predefined
subtask sequences used by the corresponding tasks.  Each trajectory is
segmented at subtask-completion events, and the resulting boundary
labels are used only for offline supervision.
Each rollout begins from a
matched task-specific initial state; object poses are randomized only
within the corresponding evaluation range.  A rollout is reset after
success, violation of a safety bound, loss of an object from the
workspace, or expiration of the task horizon.  Operator intervention
terminates the trial and is counted as a failure.

\subsection{Final-State Criteria and Safety}

For Shelf Arrangement, success requires the cola, cup, and block to be
stable in the left, center, and right shelf compartments,
respectively, without a drop or placement beyond a compartment edge.
For Drawer Restocking, the block and cup must be fully contained in the
top and middle drawers, respectively, and both drawers must finish with
less than 2\,cm residual opening.  For Tea Preparation, the system must
verify the empty top drawer, retrieve the tea bag from the middle
drawer, place it in the cup, transfer the prescribed amount of white
plastic beads from the bottle into the cup without substantial
scattering outside the cup, return the bottle to the table, and leave
the prepared cup stably on the serving tray.  The transfer outcome is
checked against the same predefined target level in every trial.

No liquid is used in the physical evaluation.  The pouring task uses a
small quantity of white plastic beads.  This dry surrogate avoids liquid
contact with the equipment and permits consistent resets while retaining
the target-directed transfer operation.  A shallow containment tray
collects any scattered beads.  Robot speed and gripper force are capped,
and drawers have mechanical travel limits.  A physical
emergency stop remains accessible; any intervention terminates and
fails the rollout.  The safety mechanisms operate independently of
SparkVLA's task-state estimation.

\end{document}